\documentclass[conference]{IEEEtran}
\IEEEoverridecommandlockouts
\usepackage{cite}
\usepackage{amsmath,amssymb,amsfonts}
\usepackage{algorithmic}
\usepackage{graphicx}
\usepackage{textcomp}
\usepackage{xcolor}
\usepackage{booktabs}
\usepackage{hyperref}
\def\BibTeX{{\rm B\kern-.05em{\sc i\kern-.025em b}\kern-.08em
    T\kern-.1667em\lower.7ex\hbox{E}\kern-.125emX}}
\begin{document}

\title{Training-Free Guidance for Discrete Diffusion Models for Molecular Generation
\thanks{This work was supported in part by the NSF under Grant 2212325. GPT 4o and GPT 4o mini were used to aid in clarity and grammar.}
}

\author{\IEEEauthorblockN{%1\textsuperscript{st}
Thomas J. Kerby}
\IEEEauthorblockA{\textit{Department of Mathematics and Statistics} \\
\textit{Utah State University}\\
Logan, USA \\
thomas.kerby@usu.edu}
\and
\IEEEauthorblockN{%2\textsuperscript{nd} 
Kevin R. Moon}
\IEEEauthorblockA{\textit{Department of Mathematics and Statistics} \\
\textit{Utah State University}\\
Logan, USA \\
kevin.moon@usu.edu}
}

\maketitle

\begin{abstract}
Training-free guidance methods for continuous data have seen an explosion of interest due to the fact that they enable foundation diffusion models to be paired with interchangable guidance models. Currently, equivalent guidance methods for discrete diffusion models are unknown. We present a framework for applying training-free guidance to discrete data and demonstrate its utility on molecular graph generation tasks using the discrete diffusion model architecture of DiGress. We pair this model with guidance functions that return the proportion of heavy atoms that are a specific atom type and the molecular weight of the heavy atoms and demonstrate our method's ability to guide the data generation.
\end{abstract}


\section{Introduction}
\label{sec:intro}

Diffusion models are a powerful method for generating data from a given distribution. To enhance their utility, extensive research has focused on developing techniques for guiding the output rather than relying solely on unconditional generation. These guidance methods have significantly evolved since the introduction of classifier guidance \cite{dhariwal_classifier_guidance}, which established diffusion models as the state-of-the-art for image generation, surpassing GANs. Classifier guidance introduced a framework allowing the gradients from a classifier to influence the generation process. The main limitation of this method is that it requires the classifier to perform well on the data at all timesteps $t$. This necessitates training a guidance model with a specific noise scheduler to augment the training data across all timesteps.

Classifier-free guidance addresses this issue by training a diffusion model that can condition on specific attributes, allowing for both unconditional and conditional generation~\cite{ho2021classifierfree}. During sampling, classifier-free guidance combines the outputs of a single model, both when conditioned on specific attributes and when unconditioned, effectively guiding the generation process toward the desired attributes, similar to how a Bayes classifier influences predictions. However, the main drawback of this method is that the attributes for guidance must be fixed beforehand during training.

A new flexible approach for guiding the sampling of diffusion models is training-free guidance, which allows guidance models to be paired with a diffusion model without requiring the guidance model to be trained on noisy data generated by the noise scheduler~\cite{univ_guide, FreeDoM, he2024manifold, shen2024training_free}. This enables the creation of foundation diffusion models that can be combined with guidance models in a plug-and-play manner. It also simplifies benchmarking and building on other researchers' work, as separate guidance models are unnecessary and can be easily shared. These improvements have made diffusion models the preferred architecture in many domains where control and human feedback are essential. Moreover, unlike autoregressive models, diffusion models learn the joint data distribution directly and do not rely on chaining conditional distributions. Their iterative nature also provides unique opportunities to guide the generation process in ways that autoregressive or purely conditional models cannot replicate.

Classifier guidance~\cite{vignac2023digress} and classifier-free guidance~\cite{ninniri2023cfreegress} have previously been implemented for graph generation using a discrete diffusion model. However, training-free guidance has not yet been extended to discrete diffusion despite its many benefits. In this paper, we introduce a framework for training-free guidance in discrete diffusion models for graph generation. We then demonstrate its effectiveness in molecular graph generation by influencing generated molecules to have a specific percentage of a given atom-type and a target molecular weight for the heavy atoms.

\section{Method}
\label{sec:background}

\subsection{Molecule Generation With Discrete Diffusion}
\label{ssec:discrete_diffusion}

We use the following notation. A molecular graph with $n$ atoms is represented as a tuple $G = (\mathbf{X}, \mathbf{E})$ where $\mathbf{X} \in \mathbb{R}^{n \times a}$ is a matrix of nodes. The $i$-th row, denoted $\mathbf{x_i} \in \mathbb{R}^{a}$, is the one-hot encoding of the atomic type of the $i$-th atom. The set of possible node types is represented by $\mathcal{X}$, with cardinality $a$. Similarly, $\mathbf{E} \in \mathbb{R}^{n \times n \times b}$ is an adjacency tensor that represents the connectivity and edge types (including the absence of a bond) where the set of possible edge types is represented by $\mathcal{E}$, with cardinality $b$. The $(i, j)$-th row of $\mathbf{E}$ , denoted $\mathbf{e_{ij}} \in \mathbb{R}^b$ contains the one-hot encoding of the bond type.

Our method is based on the graph generation model known as DiGress~\cite{vignac2023digress}, which is based on the framework introduced by~\cite{austin2021}. DiGress is a discrete diffusion model consisting of an untrained forward process $q(x^t|x^{t-1})$ and a parameterized reverse process $p_\theta(x^{t-1} | x^t)$. The forward process gradually transforms data $x^0$ into some prior (noisy) distribution $x^T$ over $T$ timesteps. This allows us to add noise to a molecular graph $G$ at time-step $t$ as follows:
\begin{align}
    q(G^t|G^{t-1}) = (\mathbf{X}^{t-1}\mathbf{Q}^{t}_X, \mathbf{E}^{t-1}\mathbf{Q}^{t}_E),
\end{align}
where $\mathbf{Q}^t_X \in \mathbb{R}^{a \times a}$ and $\mathbf{Q}^t_E \in \mathbb{R}^{b \times b}$ are transition matrices and $\mathbf{X}^{t-1}$ and $\mathbf{E}^{t-1}$ are respectively the noised node matrix and edge tensor at time step $t-1$. The structure of $\mathbf{Q}^t_X$ is:
\begin{align}
    \mathbf{Q}^t_X = \alpha^t\mathbf{I} + \beta^t\mathbf{1}_a\mathbf{m_X},
\end{align}
where $\mathbf{m_X}$ is a vector containing the marginal distribution of the node types in the training set, $\alpha^t$ is the noise scheduler, and $\beta^t = 1 - \alpha^t$. $\mathbf{Q}^t_E$ is constructed similarly, using the marginal distribution of the edge types. 

We can sample $G^t$ given $G_0$ for any $t$ in a single step:
\begin{align}
    q(G^t|G^0) = (\mathbf{X^0\Bar{Q}}_X^t, \mathbf{E^0\Bar{Q}}_E^t),
\end{align}
where $\mathbf{\Bar{Q}}_X^t = \Bar{\alpha}^t\mathbf{I} + \Bar{\beta}^t\mathbf{1}_a\mathbf{m_X}$, $\Bar{\alpha}^t = \prod_{\tau = 1}^t\alpha^\tau$, and $\Bar{\beta}^t = 1 - \Bar{\alpha}^t$. $\mathbf{\Bar{Q}}_E^t$ is constructed in the same way.

DiGress learns the parameters $\theta$ for the reverse process by optimizing the cross-entropy loss $l$ between the predicted probabilities $\hat{p}^G = (\hat{p}^X, \hat{p}^E)$ for each node and edge and the true graph $G$:
\begin{align}
    l(\hat{p}^G, G) = \sum_{1 \le i \le n}l(x_i, \hat{p}_i^X) + \gamma \sum_{1 \le i,j \le n} l(e_{ij}, \hat{p}_{ij}^E),
\end{align}
where $\gamma \in \mathbb{R}^+$ controls how much attention is paid to nodes vs the edges. This models the distribution as an independent product over the nodes and edges:
\begin{align}
    p_\theta(G^{t-1} | G^t) = \prod_{1 \le i \le n}p_\theta(x_i^{t-1}|G^t) \prod_{1 \le i,j \le n}p_\theta(e_{ij}^{t-1}|G^t).
\end{align}

To compute each term, DiGress marginalizes over the network predictions for possible node and edge attributes:
\begin{align}
    p_\theta(x_i^{t-1}|G^t) &= \int_{x_i}p_\theta(x_i^{t-1}|G^t)dp_\theta(x_i|G^t) \\
    &= \sum_{x \in \mathcal{X}}p_\theta(x_i^{t-1}|x_i = x,G^t)\hat{p}_i^X(x),
\end{align}
where
\begin{align*}
    p_\theta(x_i^{t-1}|x_i = x,G^t) = \\
    \begin{cases}
        q(x_i^{t-1}|x_i = x, x_i^t) & \text{if } q(x_i^t|x_i=x) > 0 \\
        0 & \text{otherwise}
    \end{cases}.
\end{align*}
Similarly,
\begin{align*}
    p_\theta(e_{ij}^{t-1}|e_{ij}^t) = \sum_{e \in \mathcal{E}}p_\theta(e_{ij}^{t-1}|e_{ij}=e, e_{ij}^t)\hat{p}_{ij}^E(e).
\end{align*}
% Similarly, $p_\theta(e_{ij}^{t-1}|e_{ij}^t) = \sum_{e \in \mathcal{E}}p_\theta(e_{ij}^{t-1}|e_{ij}=e, e_{ij}^t)\hat{p}_{ij}^E(e)$.
These distributions are used to sample a discrete $G^{t-1}$ from $G^t$. To sample a new graph you first sample from the prior distribution based on the training data's marginal distributions to obtain a noise graph. This is then iteratively passed through $p_\theta$ $T$ times to obtain a sampled graph.

\subsection{Training-Free Guidance for Continuous Diffusion Models}
\label{ssec:training-free_guidance}
The goal of training-free guidance is to create a conditional diffusion model from an unconditional diffusion model by pairing it with a guidance function (often also a neural network) without additional training. This allows the focus of training strong foundation diffusion models to be paired with smaller guidance models that can be tailored to a specific task. In contrast, using classifier or classifier-free guidance to target a new property to guide the generation process would require training either a new guidance model (which has access to the diffusion model's noise scheduler) or a new conditional diffusion model, respectively.

Training-free guidance models a conditional diffusion model by leveraging Bayes Rule:
\begin{align*}
    p(x_t|y) = \frac{p(y|x_t)p(x_t)}{p(y)}.
\end{align*}
Here $x_t$ is a noised latent variable at time $t$ and $y$ is some attribute we wish to condition on. When paired with taking the gradient of the log probabilities with respect to $x_t$, this produces the following:
\begin{align}
    \nabla_{x_t}\log p(x_t|y) = \nabla_{x_t}\log p(y|x_t) + \nabla_{x_t}\log p(x_t).
\end{align}
We can approximate $\nabla_{x_t}\log p(x_t)$ using the diffusion model. All that remains to create this conditional diffusion model is to model $\nabla_{x_t}\log p(y|x_t)$. 

One popular way of modeling $\nabla_{x_t}\log p(y|x_t)$ is by using an energy function. In practice, the energy function is typically approximated using a loss such as the mean-squared error between a function $f$ (often parameterized with parameters $\phi$) and some target $y$. One such example would be a function $f$ that predicts a molecular graph's drug likeliness, with $y$ being the target value. With this setup we can now model $\nabla_{x_t}\log p(y|x_t)$ with:
\begin{align}
    \nabla_{x_t}\log p(y|x_t) = -\lambda_t\nabla_{x_t}loss(f_{\phi}(\mathbb{E}_{p(x_0 | x_t)}[x_0]), y).
\end{align}

For continuous data using Gaussian noise we can use Tweedie's formula \cite{efron2011tweedie} to write $\mathbb{E}_{p(x_0 | x_t)}(x_0) = \frac{x_t - \sigma_t\epsilon_\theta(x_t, t)}{\sqrt{\alpha_t}}$ where $x_t$ represents the data at timestep $t$ and $x_t = \sqrt{\alpha_t}x_0 + \sigma_t\epsilon_t$, where $\alpha_t \in [0,1]$ decreases monotonically with $t$, $\sigma_t = \sqrt{1 - \alpha_t}$, and $\epsilon_t \sim \mathcal{N}(0, \mathbf{I})$ is random Gaussian noise. Using this we can write:
\begin{align}
    \nabla_{x_t}\log p(y|x_t) = -\lambda_t\nabla_{x_t}loss\left(f_{\phi}\left(\frac{x_t - \sigma_t\epsilon_\theta(x_t, t)}{\sqrt{\alpha_t}}\right), y\right).
\end{align}
This formula allows pre-trained networks designed for clean data to be used for the guidance process.

\subsection{Training-Free Guidance for Discrete Diffusion Models}
\label{sec:method}

The training-free guidance described in Section \ref{ssec:training-free_guidance} assumes we have a continuous $x_t$ with a noise scheduler that injects Gaussian noise. For the discrete case, we instead assume that $x_t$ is drawn from a multinomial distribution. Thus we cannot use Tweedie's formula. Instead, we can use the diffusion model $p_\theta$ such that $\mathbb{E}_{p(x_0 | x_t)}(x_0) \approx \hat{x_0}: = p_\theta(x_0 | x_t)$ if it has learned the data distribution sufficiently well. In that case, we can guide the outputs using this equation:
\begin{align}
\label{eq:dis_tfg}
    \nabla_{x_t}\log p(y|x_t) \approx -\lambda_t\nabla_{x_t}loss(f_{\phi}(\hat{x_0}), y).
\end{align}
Since $\hat{x_0}$ must satisfy the constraints of a probability distribution, we restrict the values to be non-negative and normalize the new guided version of $x_t$ to sum to 1.

To apply training-free guidance we need a function that predicts or computes a value given a graph $G$ represented as the tuple $(\mathbf{X}, \mathbf{E})$. Any function that satisfies this constraint, and preferably is not computationally expensive (since the function will be applied during all denoising steps), will work well.

The functions we consider require a sampled graph as input. However, the output of the denoising steps in our diffusion model is a probability distribution. Thus we approximate Eq. \ref{eq:dis_tfg} with:
\begin{align}
    -\lambda_t\nabla_{x_t}l(f_{\phi}(p_\theta(x_0 | x_t)), y) \approx \frac{-\lambda_t}{n}\sum_{i = 1}^n\nabla_{x_t}l(f_{\phi}(\hat{G_0^i}), y),
\end{align}
where $\hat{G_0^i}$ is sampled from $p_\theta(x_0 | x_t)$. We found in our experiments that a single sample from $p_\theta(x_0 | x_t)$ (i.e. $n=1$) is generally sufficient to obtain strong results, and as such all reported results use just one sample for estimating $\hat{x_0}$. 

\section{Experiments}
\label{sec:results}
To demonstrate our approach to training-free guidance, we trained a discrete diffusion model for molecular generation. For the diffusion model, we trained a DiGress model from the code repo provided by the DiGress authors since a pretrained version is not available for download. We trained the model on the QM9 dataset \cite{ramakrishnan2014quantum} with the main parameters specified according to the config file available on their github at \url{https://github.com/cvignac/DiGress}.

\subsection{Node Attribute Guidance}
\label{ssec:node_guidance_functions}

\begin{table}[t]
    \caption{Carbon Guidance Results - Generation results when using a function that guides the percentage of heavy atoms that are carbon atoms. We demonstrate with two target values: all heavy atoms are carbons (target $=1$) and no heavy atoms are carbons (target $=0$). For each setup 1,024 molecules were generated. Using a value of $\lambda=100,000$, the generator is able to achieve the target values for all samples.}
    \label{tab:carbon_results}
    \begin{center}
    \begin{small}
    \begin{sc}
    \begin{tabular}{rrrr}
    \toprule
    Target & $\lambda$ & \% Carbon Atoms & \% Valid \\
    \midrule
     1 & 0.0 & 0.76 +/- 0.13 & 98.3 \\
      % & 0.1 & 0.76 +/- 0.14 & 98.7 \\
      & 1.0 & 0.77 +/- 0.13 & 99.3 \\
      & 10.0 & 0.83 +/- 0.13 & 98.0 \\
      & 100.0 & 0.96 +/- 0.07 & 99.3 \\
      & 1000.0 & 0.99 +/- 0.03 & 99.7 \\
      & 10000.0 & 1.00 +/- 0.02 & 99.6 \\
      & 100000.0 & 1.00 +/- 0.00 & 99.8 \\
     \midrule
     0 & 0.0 & 0.76 +/- 0.14 & 98.3 \\
      % & 0.1 & 0.75 +/- 0.13 & 99.1 \\
      & 1.0 & 0.73 +/- 0.14 & 97.5 \\
      & 10.0 & 0.56 +/- 0.17 & 92.7 \\
      & 100.0 & 0.36 +/- 0.14 & 73.4 \\
      & 1,000.0 & 0.24 +/- 0.11 & 45.0 \\
      & 10,000.0 & 0.07 +/- 0.08 & 15.2 \\
      & 100,000.0 & 0.00 +/- 0.00 & 9.7 \\
    \bottomrule
    \end{tabular}
    \end{sc}
    \end{small}
    \end{center}
\end{table}
\begin{figure}
    \centering
    \includegraphics[width=1.0\linewidth]{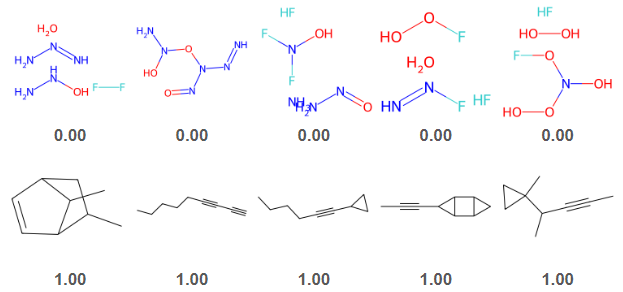}
    \caption{Examples of generated molecules using attribute guidance. The top row shows 5 uncurated samples from the 99 valid molecular graphs generated where the target proportion of heavy atoms that are carbon is $0.0$ and $\lambda=100,000$. The bottom row shows 5 uncurated samples from the 1,022 valid molecular graphs generated where the target proportion of heavy atoms that are carbon is $1.0$ and $\lambda=100,000$. At this high value of $\lambda$, the generated molecules match the target proportions exactly. However, for a target proportion of $0.0$, the validity of the generated molecules decreases as $\lambda$ increases, since pushing the carbon proportion to this extreme drives the molecules off the data manifold.}
    \label{fig:tfg_carbon}
\end{figure}

The first function that we use to guide the generation process computes the proportion of different node types. If we then compute the mean squared error between this computed proportion and some target proportion we can drive the generation process to produce molecules with the target proportion of atom types. To determine the validity of a molecular graph we use rdkit \cite{Landrum2016RDKit} and deem a graph valid if we are able to create a molecule object and successfully sanitize it.

The results for a simple demonstration of this are given in Table \ref{tab:carbon_results}, where molecules are pushed for the heavy atoms in the generated molecule to either be entirely composed of carbon atoms (target $=1$) or be anything but a carbon atom (target $=0$). In this table we see that as the $\lambda$ values increase in size, the faithfulness to the target increases, until by $\lambda = 100,000$ all 1,024 generated molecules match the target exactly. We also see that as fewer carbon atoms are included, the model is less likely to generate a valid molecule. This is expected as there are relatively few molecules without carbon atoms in the training dataset. 

% In Figure \ref{fig:tfg_carbon} we present an uncurated sample of molecules generated using this guidance function. The original DiGress model occasionally produces fragmented molecules, and our trained discrete diffusion model exhibits similar behavior. It appears that the guidance can push the model off the data manifold, increasing the frequency of these fragmented structures. This seems particularly pronounced for the target value of $0.0$ which is a harder task given that the there are few molecules in the training data without any carbon atoms. Future work will explore how to mitigate this effect.

Figure \ref{fig:tfg_carbon} shows an uncurated sample of molecules generated using this guidance function. Similar to the original DiGress model, our trained discrete diffusion model occasionally produces fragmented molecules. The guidance function appears to exacerbate this issue by pushing the model off the data manifold, increasing the frequency of fragmented structures. This effect is particularly noticeable when the target proportion of carbon atoms is $0.0$, a more challenging task due to the scarcity of carbon-free molecules in the training data. Future work will focus on mitigating this effect.

\subsection{Molecular Weight Guidance}
\label{ssec:mw_guidance_functions}

Another simple function that satisfies the desired qualities for a guidance function is the weight of a molecule's heavy atoms. We can easily calculate this at any timestep $t$ by simply summing the individual weights for each node in the graph. To accomplish this, we construct a function that multiplies the one-hot encodings of the node types for a given graph by molecular weights for each atom type and sum them up to get the molecular weight for the heavy atoms in a given graph. Using this function to guide the sampling process produces the results seen in Table \ref{tab:mw_results}. 

In this table, we observe that the sampled molecules closely match the target molecular weights while maintaining a high percentage of valid molecules, even at relatively high values of $\lambda$. The difference in the scales of $\lambda$ between Table \ref{tab:carbon_results} and Table \ref{tab:mw_results} can be explained by the difference in the scales of the loss functions. In Table \ref{tab:carbon_results}, the loss measures the difference between two proportions, whereas in Table \ref{tab:mw_results}, it measures the difference between two integers whose average is about 112. Consequently, the magnitude of $\lambda$ required to effectively guide the outputs is inversely related to the scale of the guidance loss.

\begin{figure}
    \centering
    \includegraphics[width=1.0\linewidth]{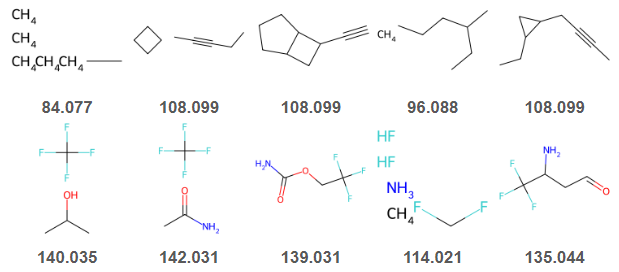}
    \caption{Examples of generated molecules using molecular weight guidance. The top row shows 5 uncurated samples from the 1,018 valid molecular graphs generated when the target weight of the heavy atoms is $105$ and $\lambda = 0.2$. The bottom row shows 5 uncurated samples from the 856 valid molecular graphs generated when the target weight of the heavy atoms is $135$ and $\lambda=0.2$.}
    \label{fig:tfg_mw}
\end{figure}

In Figure \ref{fig:tfg_mw}, we present an uncurated sample of molecules generated using this guidance function. Carbon is the lightest, and fluorine is the heaviest of the heavy atoms in the training set. Therefore, it is unsurprising that when the target molecular weight is 135, many fluorine atoms are present, despite their minority in the training data. Conversely, when the target weight is 105, we observe a higher-than-usual presence of carbon atoms.

\begin{table}[t]
        \caption{Molecular Weight Results - Generation results when guiding with the ground truth molecular weight function. For each setup 1,024 molecules were generated. As we increase $\lambda$, our model is better able to match the target weights.}
        \label{tab:mw_results}
        \begin{center}
        \begin{small}
        \begin{sc}
        \begin{tabular}{rlrr}
        \toprule
        Target & $\lambda$ & Molecular Weight & \% Valid \\
        \midrule
         105 & 0.0 & 112.41 +/- 8.86 & 98.8 \\
          & 0.0001 & 112.47 +/- 8.59 & 98.7 \\
          & 0.001 & 111.97 +/- 7.44 & 99.0 \\
          & 0.01 & 108.60 +/- 6.50 & 98.9 \\
          & 0.1 & 107.03 +/- 6.77 & 99.5 \\
          & 0.2 & 106.62 +/- 6.74 & 99.4 \\
         \midrule
         135 & 0.0 & 112.81 +/- 8.36 & 99.0 \\
          & 0.0001 & 112.90 +/- 8.87 & 98.1 \\
          & 0.001 & 117.63 +/- 11.28 & 95.1 \\
          & 0.01 & 131.60 +/- 11.26 & 85.2 \\
          & 0.1 & 135.23 +/- 7.93 & 85.6 \\
          & 0.2 & 135.95 +/- 8.17 & 83.6 \\
        \bottomrule
        \end{tabular}
        \end{sc}
        \end{small}
        \end{center}
    \end{table}

\section{Discussion | Conclusion}
\label{sec:conclusion}
In this paper, we proposed an approach for performing training-free guidance in discrete diffusion models. We demonstrated our approach using a molecular generation model, where we were able to successfully guide the model to achieve the chosen target while generating high percentages of valid molecules when generating within the data distribution. To the best of our knowledge, this is the first time that training-free guidance has been applied in discrete diffusion models.

For future work we plan to use more complex guidance models to guide the generation process such as trained neural networks. Other interesting avenues would be to apply this to other pretrained discrete diffusion models. Given that a discrete diffusion model outperformed GPT2 with slightly fewer parameters for unconditional text generation \cite{lou2024}, demonstrating the power of training-free guidance for discrete diffusion could further boost interest in using discrete diffusion models for text generation over autoregressive models. 

One limitation of our approach to training-free guidance is that it relies on the diffusion model accurately learning the underlying data distribution. In the continuous setting, training-free guidance often uses Tweedie’s formula \cite{efron2011tweedie} to efficiently estimate the original data from noisy observations, leveraging the properties of Gaussian noise injected by the noise scheduler \cite{FreeDoM, univ_guide, he2024manifold, shen2024training_free}. However, as shown in \cite{vignac2023digress}, Gaussian noise is inefficient for discrete diffusion models. Future work will explore whether analogous assumptions can be applied to discrete noise schedulers, potentially leading to more effective and efficient results.

% \cite{yang2023, dhariwal_classifier_guidance, ho2021classifierfree, univ_guide, FreeDoM, he2024manifold, klarner2024contextguided, gruver2024, ninniri2023cfreegress, chen2023edge, nisonoff2024, hoogeboom2024, austin2021, Campbell_cont_time, vignac2023digress, graph_diff_survey}

\bibliographystyle{IEEEbib}
\bibliography{main}

\end{document}